\documentclass{article}

\usepackage{arxiv}

\usepackage[utf8]{inputenc} 
\usepackage[T1]{fontenc}    
\usepackage{hyperref}       
\usepackage{url}            
\usepackage{booktabs}       
\usepackage{amsfonts}       
\usepackage{nicefrac}       
\usepackage{microtype}      
\usepackage{algpseudocode}
\usepackage{amsmath}
\usepackage{graphicx}
\usepackage{xcolor}
\usepackage{subfig}
\usepackage{csquotes}

\title{REVISITING THE DETAILS WHEN EVALUATING A VISUAL TRACKER}\title{REVISITING THE DETAILS WHEN EVALUATING A VISUAL TRACKER}

\author{
    Zan Huang \thanks{pls ping me using contact info on \url{https://github.com/pmixer} if interested in turning this technical report into a real paper.}\\
    Somewhere on Earth \\
    \texttt{huangzan@gatech.edu}
}

\begin{document}
\maketitle

\begin{abstract}
Visual tracking algorithms are naturally adopted in various applications, there have been several benchmarks and many tracking algorithms, more expected to appear in the future. In this report, I focus on single object tracking and revisit the details of tracker evaluation based on widely used OTB\cite{otb} benchmark by introducing a simpler, accurate, and extensible method for tracker evaluation and comparison. Experimental results suggest that there may not be an absolute winner among tracking algorithms. We have to perform detailed analysis to select suitable trackers for use cases.
\end{abstract}


\section{INTRODUCTION}
\label{sec:intro}

\begin{displayquote}
Everything should be made as simple as possible, but not simpler. -\emph{Albert Einstein}
\end{displayquote}

This is part of my M.Eng. thesis work in which I focused on deep learning enabled single object tracking algorithms and identified some hidden details in OTB\cite{otb} based tracker evaluation. The content is bit deprecated as the work was done more than three years ago. I just want to share the idea as these details may have not been fully explained.

In short, I show that the most commonly used area-under-curve(AUC) score of success plots in OTB\cite{otb} could be simplified to average-overlap-ratio(AOR) between tracker outputs and ground-truth bounding boxes. Based on AOR, we can design more succinct, accurate, and flexible tracker evaluation approach for more detailed analysis, to better address important features like robustness and select proper trackers for different applications.

There is no absolute winner when comparing selected representative deep trackers in this report. As we will see later, there may be an upper bound for general AOR, but when we zoom in to compare AOR of a certain tracker on different tasks, the performance changes drastically. Even trackers of similar general performance, measured by overall AOR or AUC score, would behave quite differently for the same task. That shows the diversity of trackers.

The report is for researchers and engineers who are working on visual tracking algorithms. The goal is to introduce AOR based method for tracker evaluation, prove the correctness and show the advantages. Also, due to the identified performance instability issue and diversity of visual trackers, I call for elaborating in tracker evaluation and comparison by fine-grained AOR based analysis, and considering combining distinct trackers for more robust visual tracking.

\section{METRICS REVISITED}

\begin{figure}[htbp]%
    \centering
    \subfloat[][\centering Example Precision Plot.]{\includegraphics[width=7cm]{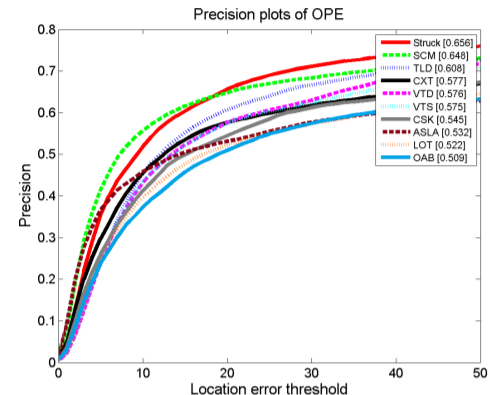} }%
    \qquad
    \subfloat[][\centering Example Success Plot.]{\includegraphics[width=7cm]{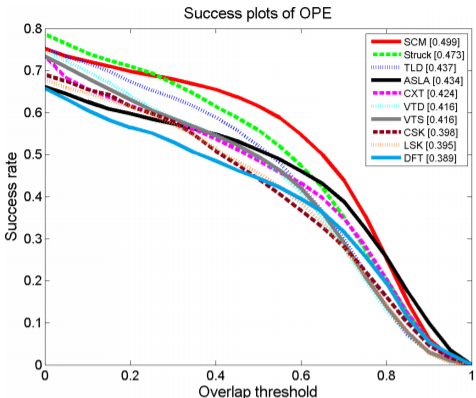} }%
    \caption{Example Plots used in OTB\cite{otb}.}%
    \label{fig:plots}%
\end{figure}

Fig.\ref{fig:plots} shows the example of precision plot and success plot used in OTB\cite{otb}, the precision plot is mainly used for comparing fixed scale trackers by only considering the box center distance between groundtruth and tracker output, the success plot AUC is more commonly used as the dominant factor for comparing trackers. In success plot, the x-axis set the threshold of groundtruth and tracker output bounding box overlap ratio, the y-axis show the percentage of success cases/frames filtered by corresponding threshold. AUC scores are computed according to the plot as the metric for measuring the performance of the tracker, higher AUC score often stands for better tracking performance. 

\begin{figure}
  \centering
  \subfloat{\includegraphics[width=9cm]{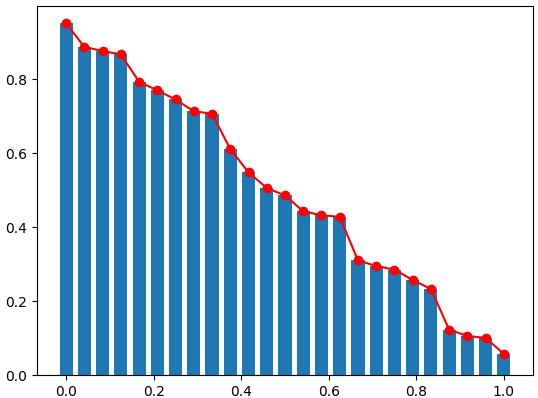}}
  \caption{Example Chart of Descendingly Sorted Overlap Ratio.}
  \label{fig:auc2aor}
\end{figure}

In this section, I show that AUC score is actually a bit inaccurate approximate of AOR on the whole testing dataset. As shown in Fig.\ref{fig:auc2aor}, if we map overlap ratios to columns of fixed width, use column height to represent the ratios and get them descendingly sorted(if non-overlap results exist for some frames, they would have overlap ratio of 0 and invisible on such bar chart).  We know that the curves shown on success plot are just smoothed line chart based on finite overlap ratio data points, and x-axis on Fig.\ref{fig:auc2aor} example bar chart corresponds y-axis on success plot while y-axis on the bar chart corresponds to x-axis on success plot, just interpreted differently. Based on basic calculus knowledge and in reference to Ferrers diagram\cite{ferrers}, we know that the AUC of success plot is equivalent to area of the bar chart shown in Fig.\ref{fig:auc2aor} which is AOR for the whole testing dataset.

 Moreover, due to the implicit smoothness and continuous assumption of success plots curves which are inconsistent with real scenario, especially considering sampling errors when calculating AUC on OTB\cite{otb}, I argue that the AUC score shown on success plot is an inaccurate approximate of AOR. As AUC score is the dominant factor when comparing trackers, I suggest directly using AOR for tracker evaluation, saving pages of space in your paper to show more informative stuff and make it easier to compare trackers in fine-grained level.
 
 In next section, I show how AOR based evaluation could help analysing and diagnosing trackers quantitatively which was hard to do as AUC is too coarse, curves are not informative enough and empirical analysis is sometimes subjective.

\section{PROPOSED METHOD}
\label{sec:method}

\begin{displayquote}
To measure is to know. If you cannot measure it you cannot improve it. -\emph{Lord Kelvin}
\end{displayquote}

After introducing AOR to replace AUC score, I show how AOR can be used extensively for detailed quantitative tracker analysis in this section. Video sequence level AOR could help us comparing trackers in different scenarios, which can further help us embracing the diversity of visual trackers. The generated sequence level AOR rank list can be used for comparing trackers quantitatively, by tracker-distance(TD) defined in this report.

Fine-grained and creative use of AOR for tracker analysis, from statistically prospective(like using the variance of AOR to represent the versatility or robustness of the tracker) or adversarial attack view(by analysing failure cases identified by low AOR video sequence segment, we can actively trace the root cause of such failure, and finally create more cases for testing, also trying possible fixes for the tracker) are highly encouraged but not covered in this report.

As visual object tracking benchmarks always consists of separate video sequences, I show how to use video sequence level AOR to explore selected tracker in the report and define tracker-distance(TD) based AOR rank list for measuring how different two trackers are. Meanwhile, One has the freedom to use AOR from frame level all the way to the whole dataset level for various statistically analysis.

For tracking tasks $T_1, T_2, ..., T_n$ on different video sequences where $n$ is the task number on the benchmark, we can compute $AOR_{T_1}, AOR_{T_2}, ...., AOR_{T_n}$ separately, each corresponds the average overlap ratio on given task(i.e. for the task named "Diving" on OTB\cite{otb}, we need to track the diver in a competition using corresponding video clip and given initial bounding box), as $\textit{AOR} \in [0, 1]$, the higher AOR suggests better tracking performance. After sorting the AOR descendingly, we can get the rank list showing  for which task, the tracker performs best and visa versa. Concrete rank list would be like $L = \{\textit{Diving}, \textit{Jump}, ...\}$ which means the tracker performs best on \textit{Diving} task of OTB\cite{otb}.

After getting the rank list for each tracker, we define the tracker-distance(TD) to quantitatively compare them. Assuming we have two rank lists $L_{\textit{tracker1}}, L_{\textit{tracker2}}$, in each of them, the element represent the tracking task(i.e. \emph{Diving}), if we assign numerical id to each task according to their rank in $L_{\textit{tracker1}}$, we can turn $L_{\textit{tracker1}}$ into $\{1, 2, ..., n\}$ and $L_{\textit{tracker2}}$ into something similar to $\{2, 7, ..., 1\}$. Finally, we can measure the difference of two tracker by counting the reverse pair number in transformed $L_{\textit{tracker2}}$, divided by maximum possible reverse pair number $n(n-1)$ to normalize the score.

TD is useful for comparing tracker as we believe similar trackers would have similar performance for the same task, thus despite concrete AOR value, the obtained rank list should also be similar for these tracker. Based on this intuition, I picked reverse pair number to construct a tracker similarity indicator. It do not require manually classifying and labelling video sequences(i.e., tagging video with occlusion by OCC on OTB\cite{otb}) and provides useful information for tracker comparison. TD can also help saving words when comparing trackers in your paper.

\section{EXPERIMENTAL RESULTS}
\label{sec:results}

Firstly, I show the necessarity of more detailed quantitatively tracker analysis by introducing an empirical bound of AOR on the whole testing dataset using another benchmark as hinted by Wang, the first author of SiamMask\cite{siammask}.

VOT\cite{vot} is a well known visual tracking competition, VOT2015 and VOT2016 share the same video source but remarked groundtruth bounding boxes, replaced manually labelled ones with segmentation based bounding boxes to maximize the overlap ratio between the bounding boxes and real object area. I computed the AOR of these two sets of groundtruth.

\begin{table}[htbp]
  \centering
  \caption{Average Overlap Ratio of VOT2015 and VOT2016 Groundtruth Bounding Boxes.}
  \scalebox{0.9}{

    \begin{tabular}{cccccccccc}
    \toprule
    Task&AOR&Task&AOR&Task&AOR&Task&AOR\\
    \midrule
    bolt2&0.770&bolt1&0.781&car2&0.857&car1&0.853\\
    birds2&0.559&sphere&0.732&singer3&0.853&singer2&0.769\\
    iceskater1&0.694&fish4&0.970&fish3&0.843&fish2&0.786\\
    fish1&0.766&marching&0.995&sheep&0.830&ball1&0.755\\
    ball2&0.670&basketball&0.722&matrix&0.795&soldier&0.794\\
    pedestrian2&0.630&pedestrian1&0.819&graduate&0.740&handball1&0.876\\
    book&0.843&helicopter&0.791&racing&0.788&butterfly&0.707\\
    soccer1&0.844&soccer2&0.758&handball2&0.781&godfather&0.782\\
    nature&0.738&octopus&0.741&blanket&0.726&hand&0.703\\
    tiger&0.811&traffic&0.811&rabbit&0.711&bmx&0.777\\
    motocross1&0.760&motocross2&0.726&gymnastics1&0.662&gymnastics3&0.729\\
    gymnastics2&0.727&gymnastics4&0.654&girl&0.766&bag&0.781\\
    tunnel&0.930&wiper&0.841&leaves&0.654&fernando&0.775\\
    glove&0.665&dinosaur&0.758&iceskater2&0.738&singer1&0.857\\
    crossing&0.839&shaking&0.809&road&0.723&birds1&0.901\\

    \bottomrule
    \end{tabular}%
    }
  \label{tab:vot1516overlap}%
\end{table}%

The raw AOR data is shown in Table.\ref{tab:vot1516overlap}, and overall AOR is $0.772$ for the whole dataset. Even groundtruth differs greatly from overlap prospective. I suspect that there is an empirical upper bound for overall performance of trackers on the benchmark. Instead of putting too much focus on the leaderboard, we need pay more attention to details, especially to failure cases and imbalanced performance issue for usability and robustness when general AOR is approaching $0.7$.

Secondly, with GOTURN\cite{goturn}, SiameseFC\cite{siamesefc}, CFNet\cite{cfnet}, ECO\cite{eco}, STCT\cite{stct}, MDNet\cite{mdnet} picked
for experiments, I show the imbalanced sequence level AOR issue which got ignored when we only comparing overall performance of given trackers on the whole dataset. Moreover, we can see the diversity of visual trackers by inspecting the issue.

\begin{table} [htbp]
  \caption{Partial Result of Descendingly Ranked Sequence Level Average Overlap Ratio of Selected Trackers on OTB.}
    \resizebox{\textwidth}{!}{
    \begin{tabular}{cccccccccccc}
    \toprule
    GOTURN&AOR&SiameseFC&AOR&CFNet&AOR&ECO&AOR&STCT&AOR&MDNet&AOR\\
    \midrule
\textcolor[rgb]{0,1,0}{Diving}&0.528&Human2&0.727&Walking&0.719&Girl&0.795&David3&0.753&Human5&0.735\\
\textcolor[rgb]{1,0,0}{Jump}&0.126&MotorRolling&0.345&DragonBaby&0.254&Freeman4&0.529&Skating2-1&0.469&Human3&0.515\\
BlurOwl&0.115&Human9&0.341&Lemming&0.213&Skating1&0.529&Surfer&0.447&BlurOwl&0.493\\
Board&0.102&Skating2-1&0.325&Matrix&0.209&Football1&0.513&Biker&0.424&Gym&0.475\\
Tiger1&0.101&ClifBar&0.288&Subway&0.192&Biker&0.510&DragonBaby&0.412&Vase&0.463\\
Woman&0.101&Coupon&0.240&Soccer&0.176&Skiing&0.491&ClifBar&0.389&Dog&0.461\\
Doll&0.097&Football&0.237&MotorRolling&0.164&Trans&0.485&\textcolor[rgb]{0,1,0}{Diving}&0.381&Ironman&0.444\\
BlurCar1&0.087&Soccer&0.173&Ironman&0.162&Gym&0.476&Dog&0.360&Soccer&0.415\\
Soccer&0.085&Human4-2&0.161&Human9&0.162&Skating2-1&0.475&Panda&0.300&Bird1&0.393\\
BlurCar3&0.082&Bird1&0.145&Tiger1&0.146&Ironman&0.407&\textcolor[rgb]{1,0,0}{Jump}&0.274&\textcolor[rgb]{0,1,0}{Diving}&0.380\\
Subway&0.076&\textcolor[rgb]{0,1,0}{Diving}&0.127&\textcolor[rgb]{0,1,0}{Diving}&0.108&Coupon&0.381&Freeman4&0.267&Biker&0.372\\
Human4-2&0.062&Board&0.122&Skiing&0.107&Skating2-2&0.352&Soccer&0.207&Coupon&0.338\\
Bird1&0.055&Human8&0.072&\textcolor[rgb]{1,0,0}{Jump}&0.097&\textcolor[rgb]{0,1,0}{Diving}&0.322&Matrix&0.194&Box&0.333\\
Singer2&0.051&\textcolor[rgb]{1,0,0}{Jump}&0.065&Girl2&0.065&Panda&0.317&Skiing&0.092&Matrix&0.279\\
CarDark&0.050&Singer2&0.038&Jumping&0.063&Bird1&0.180&Girl2&0.069&Football&0.140\\
Human3&0.026&Human3&0.013&Singer2&0.040&MotorRolling&0.092&Ironman&0.063&Crowds&0.094\\
Bolt&0.006&Bolt&0.011&Human3&0.008&\textcolor[rgb]{1,0,0}{Jump}&0.039&Bird1&0.030&\textcolor[rgb]{1,0,0}{Jump}&0.082\\
    \bottomrule
    \end{tabular}%
    }
  \label{tab:avgoverlapvalues}%
\end{table}%

Table.\ref{tab:avgoverlapvalues} contains partial results of rank lists for each selected tracker(only retain the best performing tasks and worse performing tasks to make the table shorter). The highest AOR is way better than worse AOR for each tracker which suggests that the performance of trackers is unstable, while these low performing tasks commonly corresponds to shorter video sequences, which means even trackers performs terrible for these tasks, it could be easily ignored due to their low weights when contributing the overall AOR or AUC score. We need to report average of video sequence level AORs when presenting general visual tracker claiming being able to handle various objects in different scenarios.

Moreover, like what is shown by colored entries in Table.\ref{tab:avgoverlapvalues}, GOTURN\cite{goturn} which generally performs worse when comparing with other trackers do have much better performance for tasks like \emph{Diving} and \emph{Jump}. When measuring the TD of selected trackers, as presented in Table.\ref{tab:trackerdist}, we can see that GOTURN\cite{goturn} is indeed more different from other trackers, the TD of GOTURN\cite{goturn} to other trackers are marginally larger than other entries.

\begin{table*}[htbp]
  \centering
  \caption{Tracker Distance($TD$) Comparison.}

    \begin{tabular}{ccccccc}
    \toprule
    &GOTURN&SiameseFC&CFNet&ECO&STCT&MDNet\\
    \midrule
    GOTURN& 0.0&  0.353&  0.421&  0.458&  0.462&  0.417\\
    \midrule
    SiameseFC& 0.353&  0.0&  0.295&  0.251&  0.301&  0.288\\
    \midrule
    CFNet& 0.421&  0.295&  0.0&  0.258&  0.252&  0.261\\
    \midrule
    ECO& 0.458&  0.251&  0.258&  0.0&  0.182&  0.220\\
    \midrule
    STCT& 0.462&  0.301&  0.252&  0.182&  0.0&  0.242\\
    \midrule
    MDNet& 0.417&  0.288&  0.261&  0.220&  0.242&  0.0\\
    \bottomrule
    \end{tabular}%
  \label{tab:trackerdist}%
\end{table*}%

As I showed the single tracker performance instability issue and the diversity of trackers, it is obviously better trying to combine distinct trackers for more robust visual tracking than just claiming some of them can well handle all tracking tasks independent of the object and scenario. But this topic is underexplored according to my limited knowledge.


\section{CONCLUSION}
\label{sec:conclusion}

OTB\cite{otb} proposed evaluation metric is still widely used, while newer benchmarks already practiced using average overlap ratio for simplicity. The report is for encouraging the use of AOR over classical OTB\cite{otb} plots with theoretical basis to increase information intensity in related papers, so we can perform deeper discussion on tracker evaluation.

AOR based tracker evaluation method is more intuitive, accurate, and flexible, supporting multi level tracker performance inspection, from single frame level all the way to the whole testing dataset level evaluation, and the generated numbers are more statistical analysis friendly, which is good for research and industry-use-oriented engineering.

I gave an empirical upper bound for overall average overlap ratio, introduced the tracker performance instability issue, showed the diversity of trackers as revealed by TD measure, identified the formerly ignored short-sequence-bad-performance problem. All supported by proposed AOR based tracker evaluation.

The idea of coarse-to-fine evaluation is not limited to visual tracking research, it could be easily extended to diagnose other kinds of models, like detectors etc. The main idea is to identify and address the imbalanced performance issue and embrace diversity of formerly proposed models. To escape the involution, take research outcomes to production.

\bibliographystyle{unsrt}  
\bibliography{main}

\end{document}